\documentclass[twoside]{article}

\usepackage{booktabs}
\usepackage[accepted]{aistats2020}
\usepackage{natbib}

\usepackage{amsmath,amsfonts,bm}

\def\eqref#1{equation~\ref{#1}}

\def\1{\bm{1}}

\def\vf{{\bm{f}}}

\def\vw{{\bm{w}}}

\DeclareMathAlphabet{\mathsfit}{\encodingdefault}{\sfdefault}{m}{sl}
\SetMathAlphabet{\mathsfit}{bold}{\encodingdefault}{\sfdefault}{bx}{n}

\newcommand{\E}{\mathbb{E}}

\newcommand{\softplus}{\zeta}

\usepackage{hyperref}
\usepackage{url}
\usepackage{graphicx}
\usepackage{multirow}
\usepackage{caption}
\usepackage{xcolor}

\begin{document}

%
\runningtitle{Learning Dynamic and Personalized Comorbidity Networks from Event Data using DDP}

%
\runningauthor{Z. Qian, A. M. Alaa, A. Bellot, J. Rashbass, M. van der Schaar}

\twocolumn[

\aistatstitle{Learning Dynamic and Personalized Comorbidity Networks \\ from Event Data using Deep Diffusion Processes}

\aistatsauthor{ Zhaozhi Qian \And Ahmed M. Alaa \And  Alexis Bellot \\ \And Jem Rashbass \And Mihaela van der Schaar}

\aistatsaddress{ University of Cambridge \And  UCLA \And University of Cambridge \\ \And NHS Digital \\ Public Health England \And University of Cambridge \\ UCLA \\ The Alan Turing Institute} ]

\begin{abstract}
Comorbid diseases co-occur and progress via complex temporal patterns that vary among individuals. In electronic health records we can observe the different diseases a patient has, but can only infer the temporal relationship between each co-morbid condition. Learning such temporal patterns from event data is crucial for understanding disease pathology and predicting prognoses. To this end, we develop {\it deep diffusion processes} (DDP) to model ``dynamic comorbidity networks'', i.e., the temporal relationships between comorbid disease onsets expressed through a dynamic {\it graph}. A DDP comprises events modelled as a multi-dimensional point process, with an intensity function parameterized by the edges of a dynamic weighted graph. The graph structure is modulated by a neural network that maps patient history to edge weights, enabling rich temporal representations for disease trajectories. The DDP parameters decouple into clinically meaningful components, which enables serving the dual purpose of accurate risk prediction and intelligible representation of disease pathology. We illustrate these features in experiments using cancer registry data.
\end{abstract}

\section{INTRODUCTION}
The illnesses arise not just from individual causes for the specific disease but as a complex interaction between other diseases the patient already had. Identifying and understanding the  contribution of comorbidities to disease progression and outcomes is fundamental to medicine and clinical practice. The causal structure of relationships between diseases can be represented by networks that are dynamic in nature. For instance, long-term diabetes increases the risk of cardiovascular and renal disease making high blood pressure and its complications - such as heart attacks more likely. The strength of edges between network nodes changes over time, depending on the entire patient history. Beyond disease progression, these dynamics are also prevalent in economics, finance and sociology \textcolor{blue}{\citep{ahmed2009recovering,namaki2011network}}. 

In most of these cases, the underlying network dynamics are unknown, and what we observe are sequences of events spreading over the network. To infer the latent  network dynamics from observed sequences, one needs to take into account both \textit{when} and \textit{what} events occured in the past since both carry information on the mechanisms involved in disease instantiation and progression. Multi-dimensional point processes are natural candidates for this problem; they explicitly model the time period between events as random variables, and allow them to modulate the \textit{intensity function} --- a stochastic model for the time of the next event given previous events. However, traditional parametric models are not expressive enough to capture network dynamics, i.e. the networks they learn are static in nature. On the other hand, existing neural point process models do not entail well-defined network structure due to their complex parameterization.

\begin{figure*}[ht]
\centering
\includegraphics[width=6in]{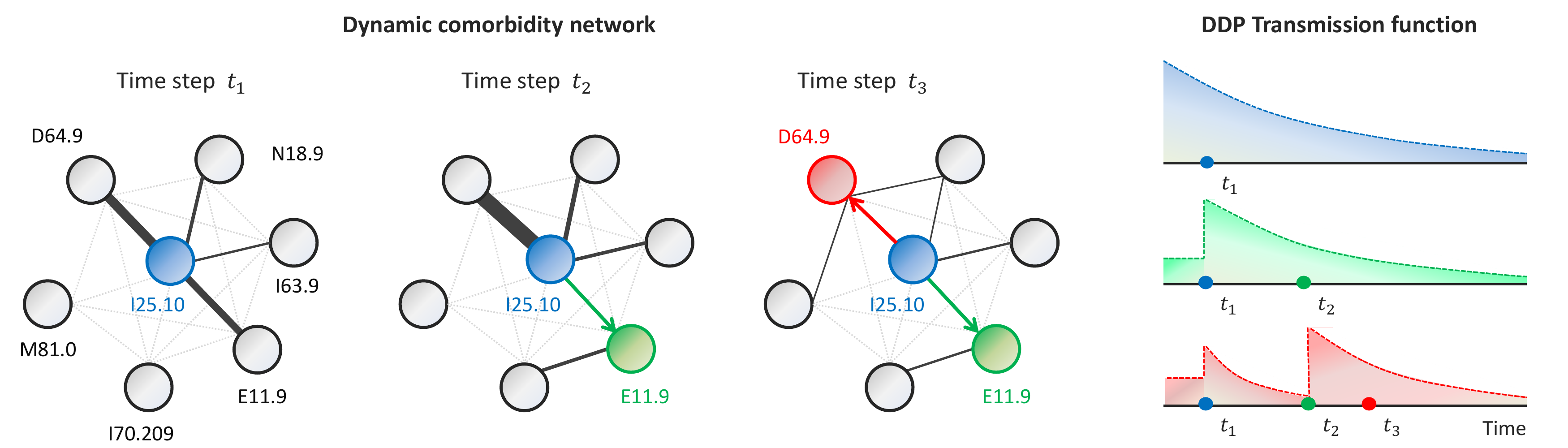}
\caption{{\footnotesize An Exemplary Realization of DDP. Each node corresponds to a disease ICD-10 code} {\scriptsize (D64.9: Anemia, N18.9: Renal failure, I63.9: Cerebral infarction, E11.9: Diabetes, I70.209: Atherosclerosis, M81.0: Osteoporosis, and I25.10: Heart disease.)} {\footnotesize Edge weights are depicted via their thickness. The left panels show the evolution
of the disease network, and the righmost panel shows the intensity functions of three selected nodes. Onset of \textcolor{blue}{heart disease} (at $t_1$) triggers spikes in the intensity functions of \textcolor{green}{diabetes} and \textcolor{red}{anemia}, making them more likely in the future. The onset of \textcolor{green}{diabetes} (at $t_2$) elevates the risk of \textcolor{red}{anemia} (i.e., thicker edge), which consequently occurs at $t_3$. Edge weights are modulated by a neural network over time.}}
\label{PP}
\end{figure*}

In this paper, we develop the {\it deep diffusion process} (DDP), a deep probabilistic model for diffusion over comorbidity networks based on mutually-interacting point processes as illustrated in Figure 1. We model the DDP intensity function as an combination of contextualized background risk and networked disease interaction. The disease interaction further consists of three components: (1) static pairwise interactions, (2) time influence, and (3) dynamic influence factors. The first two components are standard in point process models whereas the last component makes use of a deep neural network to (dynamically) update the disease's influence on future events. The introduction of neural networks does not only add to model capacity, but also enables principled predictions based on clinically interpretable parameters which map the patient history on to personalized comorbidity networks. This brings us closer to understanding the underlying disease mechanisms, which as we hypothesize, leads to better out-of-sample and out-of-domain performances. In our experiments, we provide encouraging results in this direction, with better performance of our model in medical data from a different domain.

\section{RELATED WORK} \label{sec:related_work}
In this Section, we highlight previous approaches based on point process formalism, and techniques specifically used in medicine that are relevant to our problem.

\subsection{Point Processes for Event Streams}

\subsubsection{Parametric Models}

\textcolor{blue}{\citet{gomez2012inferring}} introduced one of the earliest algorithms for discovering latent networks from sequences of events with a transmission process influenced only by the most recent event.
The \textbf{cHawkes} model \textcolor{blue}{\citep{choi2015constructing}} removes the Markovian assumption by modelling the event stream as a Hawkes Process where past events temporarily raise the probability of future events. The resulting network captures the pairwise interaction between any two events. However, the influence of the \textit{combinations} of previous events and their \textit{timing} is not accounted for in the network structure. As a result, the learnt network is constant for all event streams at all time. Hence, we refer to it as a \textit{static population-level} network.

\subsubsection{Neural Network Based Models}

Several recent publications have been focusing on expanding the flexibility of point process models by using recurrent neural networks (RNN).

\textcolor{blue}{\citet{du2016recurrent}} models the inter-arrival time between consecutive events as a univariate point process and annotates each event with a marker to indicate the event type. Importantly, the marker and the arrival time of the next event are conditionally independent given the history. The independence assumption imposes limitations on the expressiveness of the model as there is only one underlying intensity function for all types of events.
\textbf{Neural Hawkes} \textcolor{blue}{\citep{mei2017neural}} models the intensity function directly as a continuous time LSTM. The resulting model has much better flexibility and has achieved the state-of-the-art performance on a variety of prediction tasks. However, the model does not generate a well-defined network between events and it lacks interpretability in general as the hidden state of the RNN do not correspond to clinically meaningful variables. 

More recently, the \textbf{RPPN} model \textcolor{blue}{\citep{xiao2019learning}} incorporates temporal attention mechanism to improve the interpretability of neural point process. The model requires a separate attention function for each possible event type in order to connect the observed past with the unobserved future. This may not be an issue when all types of events occur relatively often, but in the medical domain, the majority of diseases have low prevalence in the population\footnote{For example, heart disease is perceived to be very common but it actually occurs in only 1.07\% of adults according to official statistics \textcolor{blue}{\citep{prevdata}}. Rare diseases often have prevalence lower than 0.01\%.}, which means the attention functions for these diseases may not be adequately trained due to scarcity of data. 
\textcolor{blue}{\citet{lamprier2019recurrent}} also considered applying neural networks to information diffusion modelling, although the model does not allow the same type of event to occur more than once. In the medical setting, recurrence of previous diseases carries important information about the patient's health condition. It is also of interest to predict the future recurrence of existing morbidities. 

\begin{table}[ht]
\label{tab:intensity_comparison}
\begin{center}
\caption{{\footnotesize Point Process intensity functions. Subscript $v$ denotes event type. $\bm{f}$ is the context vector. $\gamma$ is the time-influence kernel. $\bm{h}_t$ is the outputs of RNN units and $A_{\cdot, v}$ is the output of event $v$'s attention function. $\softplus$ and $\sigma$ are softplus and sigmoid functions respectively.}}
\resizebox{0.48\textwidth}{!} {
    \begin{tabular}{lll}
    \hline

    {\bf Model}           & {\bf Background}  & {\bf Temporal Dependence} \\ \hline
    & & \\ 
    Poisson Process & $\mu_v(t)$            & 0                    \\
    Hawkes Process  & $\mu_v(t)$            & $\sum_{i:t_i<t}\alpha_{v_i, v}\gamma_{v_i, v}$                 \\
    cHawkes           & $\mu_v(t, \bm{f})$    & Same as above                  \\
    Neural Hawkes   & 0                   & $\softplus(\vw_v \bm{h}_t)$                  \\
    RPPN            & 0                   & $\softplus(\vw \sum_{i:t_i<t} A_{v_i, v} \bm{h}_{t_i} \gamma_{v_i, v} )$                  \\ 
    DDP             & $\mu_v(t, \bm{f})$    & $\sum_{i:t_i<t}\alpha_{v_i, v}\gamma_{v_i, v}\sigma(\vw \bm{h}_t)$  \\ [1ex]
    \hline
    \end{tabular}
}
\vspace{1ex}
\end{center}
\end{table}
\vspace{-.25in}

\subsection{Medical Disease Networks}
\label{sec:med_network}
Within the medical community, understanding disease networks and associated comorbidities --- i.e. any two or more diseases that occur in one person at the same time --- is fundamental to the diagnosis and treatment of patients. Many rule-based scoring models are based on empirical association of symptoms and clinical outcomes. For example, the Charlson Comorbidity Index \textcolor{blue}{\citep{charlson1987new}} was proposed as early as 1987 to predict the ten-year mortality for a patient by summing up the risk indices associated with various comorbid conditions. The index remains the preferred approach in medical community to represent comorbidity history \textcolor{blue}{\citep{quan2011updating}}. 

Recent works have also investigated the construction~of data-driven dynamic disease networks \textcolor{blue}{ \citep{hu2019large,lee_inference_2019,beck2016diagnosis,hidalgo_dynamic_2009}}. However, with no exception, the networks in these works are constructed in two steps. First, certain pairs of diseases are linked together based on population level statistics such as risk ratio or temporal correlation. Next, the disease pairs are pieced together into longer trajectories or networks. Since all the information used in this process is on population level, the resulting graph is not \textit{personalized}. Furthermore, constructing the network by combining pieces usually implies strong independence assumptions e.g. Markovian assumption, which rarely holds in disease progression (a real example is given in the appendix). Therefore, the \textit{dynamic} aspect of disease progression is not adequately represented. 

The main contribution of this work is to augment the above approaches by modelling the disease network itself as an individualized dyanmic graph. This allows us to model more complex temporal interactions between diseases as well as provide personalized predictions.

\section{DEEP DIFFUSION PROCESS}
\subsection{Dynamic Network Representation}

Consider a dynamic network $G_t = (V, L_t, E_t, W_t)$ consisting of a set of vertices $V = \{1,...,K\}$ annotated with binary labels $L_t = \{0,1\}^K$, and a set of directed edges $E_t$ weighed by $W_t = \mathbb R^{+|E|}$. The vertices $V$ correspond to the set of all possible event types. 
At any time, a vertex $v$ has label $l_v=1$ if a type $v$ event has occurred or $l_v=0$ otherwise. The edge set, formally defined as $E \subseteq \{(v_i, v_j) | v_i, v_j \in V, l_{v_i} = 1\}$, contains edges that link an observed vertex $v_i$ to another vertex $v_j$ if $v_i$ modulates $v_j$'s chance of occurrence. 
The edge weights $W_t$ represent the strength of such modulation effect between events (Refer to Figure 1).

While we do not observe the network directly at each time, we do have access to individual trajectories through the network, available as a sequence of events and corresponding time points,
\begin{align}
    \mathcal H = \{(t_1, v_1),...,(t_n, v_n)\}
\end{align}
where $t_i \in \mathbb R^+$ is the time of occurrence and $v_i\in V$ is the associated event type. From the event sequence, one can immediately derive the vertex label $L_t$ for $t \le t_n$. Since the vertex set $V$ is fixed a-priori depending on the problem scope, the remaining unknown components of the graph are the label $L_t$ for $t > t_n$ as well as the weighted edges $E_t$, $W_t$ for $t > 0$. 

Determining the future vertex label is~known~as~\textit{event prediction}, whereas~uncovering~weighted~edges corresponds to \textit{network inference}. Our goal is to devise a model that addresses both problems simultaneously.

\subsection{Preliminaries on Point Process}

Before formally introducing the DDP model, we first recapitulate several key concepts of point process. 

Lying at the core of point processes is the {\it intensity function} $\lambda_v(t)$, which is the probability of event $v$ occurring in time window $[t, t + dt)$ given a history $\mathcal H_t := \{(t_i,v_i):t_i<t\}$, \textit{i.e.},
\begin{align}
    \lambda_v(t)dt:= Pr(\text{event of type $v$ in } [t,t+dt) | \mathcal H_t )
\end{align}
As we can see in Table 1, different point process models have different parameterizations of the intensity function ranging from the simplest Poisson process to the complex Neural Hawkes and RPPN. However, once the intensity function is given, many interesting properties can be readily derived. For example, the likelihood of an observed sequence $\mathcal H_T$ is given by
\begin{align} \label{eqn:likelihood}
    \mathcal L(\theta;\mathcal H_T) := \prod_{(t_i,v_i)\in \mathcal H_T} \lambda_{v_i}(t_i)\exp\left(-\int_{t_{i-1}}^{t_i} \lambda(\tau)d\tau\right),
\end{align}
where $\lambda(t) = \sum_{v \in \mathcal V} \lambda_{v}(t)$ and $\theta$ is the collection of all free parameters in the model.
As another example, the probability of an type $v$ event happening at a \textit{specific} time $t_{i+1} > t$ is given by
\begin{equation} \label{def:classification_proba}
P(v_{i+1}=v \,|\, \mathcal H_t, t_{i+1}) = \frac{\lambda_v(t_{i+1})}{\lambda(t)}
\end{equation}
and the occurrence time of the next event is given by
\begin{align} \label{def:next_event_time}
    P(t_{i+1} = t \,|\, \mathcal H_T) = \lambda(t) \exp\left(-\int_{t_{i}}^t \lambda(\tau)d\tau\right)
\end{align}
The model can be trained in multiple ways. In general, one can maximize the likelihood function \ref{eqn:likelihood} via stochastic gradient descent. If the integral term does not have a closed form, it can be approximated by Monte Carlo sampling as done in \textcolor{blue}{\citet{mei2017neural}}.
In addition, it is also possible to train the model by minimizing the prediction loss based on (\ref{def:classification_proba}) and (\ref{def:next_event_time}) \textcolor{blue}{\citep{du2016recurrent}}. 

\subsection{Model Specification}

This section presents the Deep Diffusion Process --- a deep probabilistic model for inferring network dynamics while accurately predicting future events.

\subsubsection{intensity function}

Our objective is to enrich the intensity function in order to capture the time-dependent disease-to-disease relationships. To this end, we decompose the overall intensity function into two additive components:
\begin{equation} \label{equ:intensity_decomp}
    \lambda_v(t) = \mu_v(\vf) + \sum_{t_i<t} g_{v}(v_i, t_i, \mathcal H_{t_i}, t).
\end{equation}
The first term captures the occurrence of events due to static exogenous risk factors $\vf$. For example, it can model the increased risk of heart attack among the obese patients. The second term models the impact from past events. Each historical event adds an ``impulse'' $g_v$ to the intensity function of event $v$ depending on the event type $v_i$, the timing $t_i$, and, most importantly, the event history $\mathcal H_{t_i}$ at the time. This decomposition allows modeling the impact of exogenous factors and that of past events separately.

Next, we introduce the parametric form to capture the impact from past events as follows:
\begin{equation} \label{equ:trigger}
    g_{v}(v_i, t_i, \mathcal H_{t_i}, t) = \alpha_{v_i, v} \cdot \gamma_{v_i, v}(t-t_i)\cdot w_i(\mathcal H_{t_i}).
\end{equation}
The parameter $\alpha_{v_i, v}$ captures the instantaneous impact from event $v_i$ to $v$. We use $\alpha$ to denote the matrix that contains all the $\alpha_{v_i, v}$ between any two events. The time influence kernel $\gamma_{v_i, v}(t-t_i)$ captures the decay or increase of previous events' influence. It is a non-negative function defined on $\mathbb R^+$ and integrates to one. One common choice is the exponential kernel $\gamma_{v_i, v}(t-t_i) = \beta_{v_i, v}\exp(-\beta_{v_i, v}(t-t_i))$. 

\begin{figure}[t]
\centering
\includegraphics[width=3.25in]{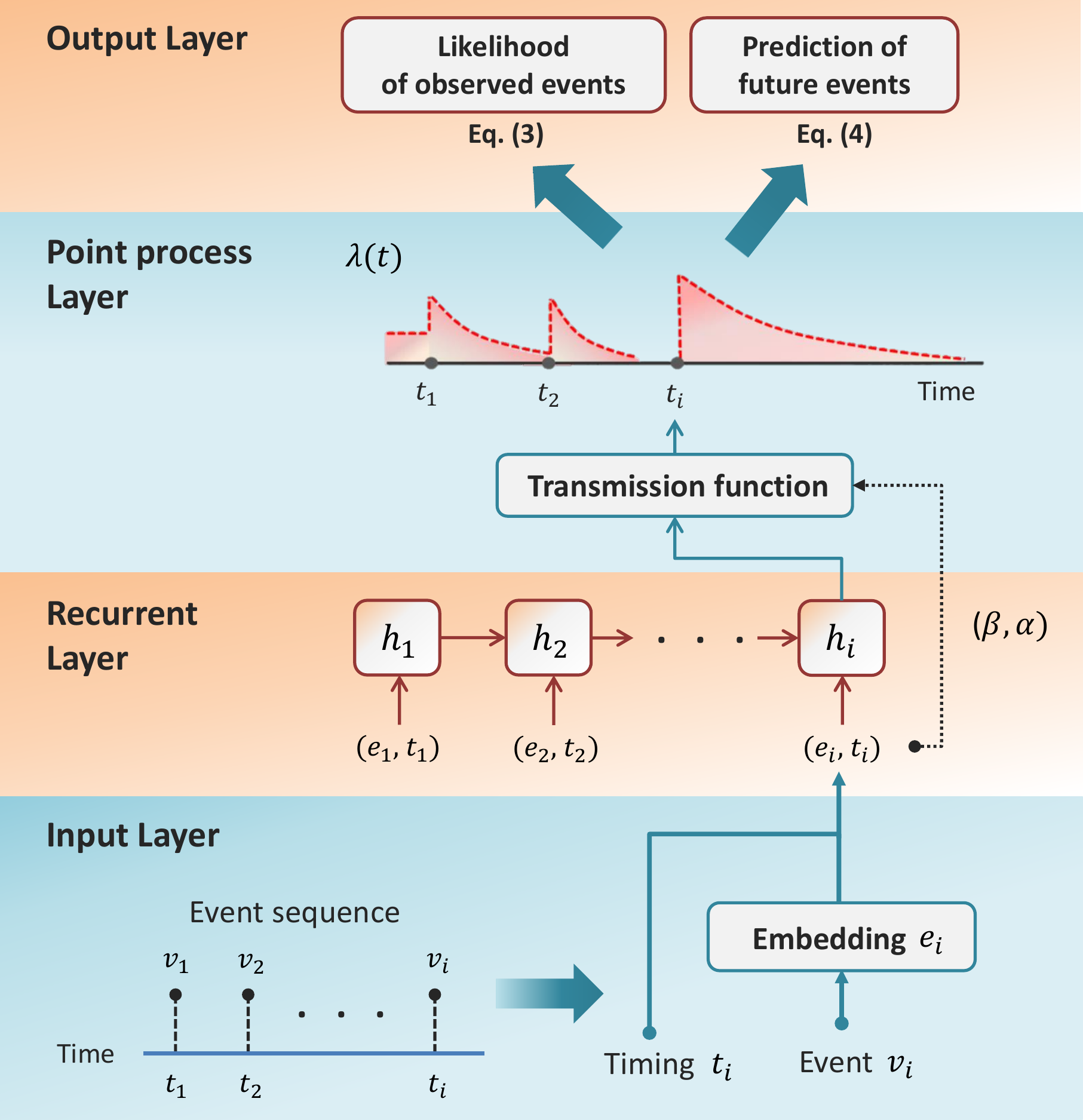}
\caption{Schematic of the DDP Architecture.}
\label{fig:archi}
\end{figure}

The last component, $w_i(\mathcal H_{t_i}) \in [0, 1]$, is the (dynamic) influence factor that depends on the full patient history. It is learned by an RNN applied to the sequence of past events. As shown in Figure \ref{fig:archi}, the event $v_i$ is encoded through an embedding $e_i \in \mathbb R^D$, which together with the time gap $\Delta t_i := t_i - t_{i-1}$ are fed into the recurrent layer as follows: 
\begin{equation} \label{equ:lstm}
    h_i = LSTM(e_i \odot \Delta t_i, h_{t-1}),
\end{equation}
where $h_i$ is the LSTM output. In our implementation, we used standard LSTM with the time gap as an additional input dimension as shown in (\ref{equ:lstm}). However, we note that any continuous-time RNN (e.g., phased LSTM \textcolor{blue}{ \citep{neil2016phased}}) is applicable. The influence factor $w_i$ is then given by
\begin{align} \label{equ:weight}
    w_i = \sigma(h_i \cdot W + b),
\end{align}
where $W$ and $b$ are parameters to be learned, and $\sigma(.)$ is the sigmoid function.   

For training, we use a loss function comprising the likelihood function $L(\mathcal H_T;\theta)$ in (\ref{eqn:likelihood}) and the cross entropy loss $l_p(\mathcal H_T;\theta)$ for event type prediction, i.e.,  
\begin{align} \label{eqn:objective_function}
    \theta^*=\operatorname{argmax} \mathcal L(\mathcal H_T;\theta) - \eta \cdot l_p(\mathcal H_T;\theta)
\end{align}
where $\eta>0$ is a hyperparameter that trade -off the two objectives, and is determined from a validation set. The loss function in (\ref{eqn:objective_function}) encapsulates our dual objective of a faithful representation for the observed event sequence and the ability to predict the next event.

\subsection{Dynamic Network Inference}

The parameter $\alpha$, the time influence kernel $\gamma$ and the influence factors $w_{i}$ jointly define the network structure at time $t$. $\mathbf\alpha$ is the baseline matrix that encodes static pairwise relationships. The larger the value of $\alpha_{uv}$, the more influence event $u$ will have on event $v$ \textit{on average}. The time influence kernel further modulates the link strength based on the time gap between the occurrence of events.

The influence factor $w_{i}$ modulates $\mathbf\alpha$ and enables the network structure to adapt to the observed event sequence. Based on the full history of past events, the influence factor may strengthen or diminish the impact of one particular event and thus modifies all its outgoing links. Therefore, at time $t$, the directed edge $v\rightarrow u$ will have weight
\begin{equation} \label{equ:edge_weight}
    W_{v\rightarrow u}(t) = \sum_{\substack{(t_i, v_i) \in \mathcal H_{T} \\ v_i = v}} \alpha_{v, u} \cdot \gamma_{v, u}(t-t_i) \cdot w_i.
\end{equation}
It is worth highlighting that the resulting graph is dynamic in two aspects. First, the influence factor $w_i$ is updated for each event $v_i$ based on the full event history up to that point. Depending on the combination and the timing of historical events, the influence factor for subsequent events will differ, leading to a different graph structure. Secondly, the time influence kernel $\gamma$ modulates the edge weight as time moves on.

It is often desirable for the graph to have a sparse structure i.e. $W_{v\rightarrow u} = 0$ for many pairs of events $v, u$. We can introduce a $L_1$ regularization term for $\alpha$ matrix to encourage sparsity as proposed in \textcolor{blue}{\citet{choi2015constructing}}.

Lastly, we note that sometimes it is required to construct a population-level static graph instead of a dynamic graph to capture the high-level event interaction. Static edge weights can be found by averaging out the dynamic components in equation \ref{equ:edge_weight} as follows
\begin{equation} \label{equ:edge_weight_static}
    W_{v\rightarrow u} = \alpha_{v, u} \E[w_i],
\end{equation}
where the expectation represents the average influence factor of event $v$. 

\section{EXPERIMENTS}
In this Section, we utilize data from a large-scale cancer registry to evaluate DDP\footnote{Implementation details are provided in the appendix}. Throughout our experiments, we evaluate DDP with respect to three aspects: (a) its ability to extract interpretable disease networks that are sensible in the light of current medical literature (Section 4.2), (b) its accuracy in predicting disease pathways (Section 4.3), and (c) its generalizability to out-of-domain datasets (Section 4.4).  

\subsection{Data Description}

We used national registry data for a cohort of colorectal cancer patients diagnosed between 2011 and 2015. The data comprises 268,000 observations of 100 common diagnoses for 54,000 patients. Each patient is associated with up to 15 comorbidities. 
The earliest diagnoses date back to the 1990s, which gives us a fairly broad timescale to study the progression of colorectal cancer. 
In addition to the primary dataset described above, we have also considered data for 25,000 patients with stomach cancer. It is well understood that patients with stomach cancer are exposed to different risk factors from patients with colorectal cancer \textcolor{blue}{\citep{miller1982risk, drasar1973environmental}}. Therefore, we use this dataset as an \textbf{out-of-domain} test set to validate transferability and robustness of DDP. 

\subsection{Colorectal cancer comorbidity networks}

\begin{figure*}[t]
\centering
\includegraphics[width=6.5in]{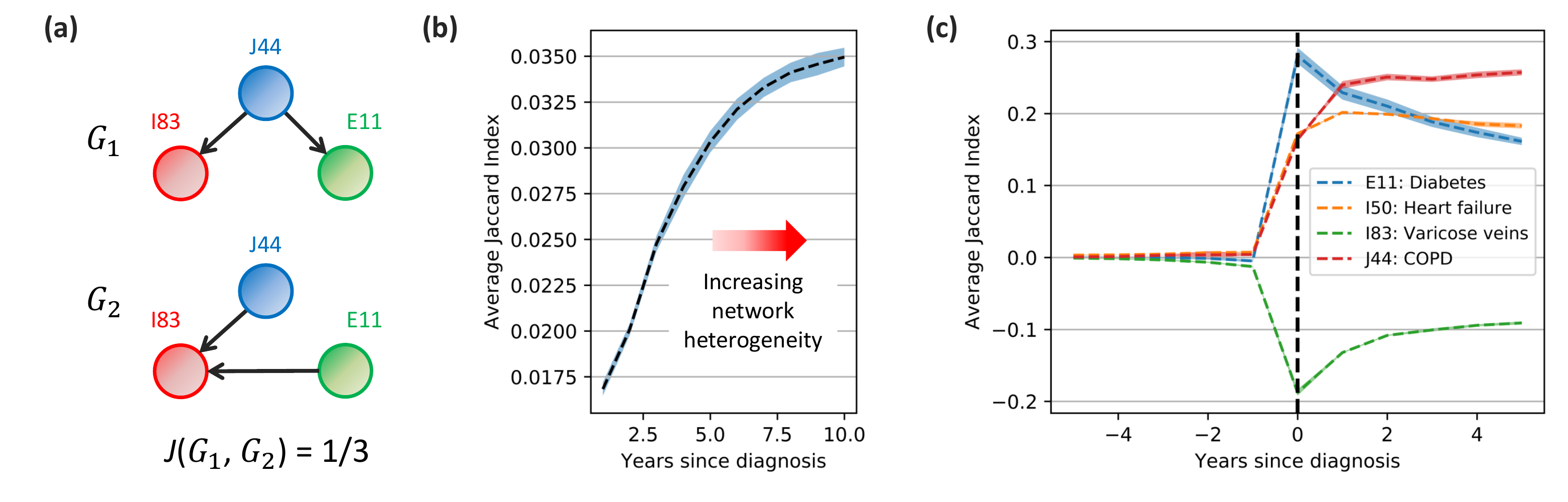}
\caption{(a) Illustration of the Jaccard index between two exemplary graphs. (b) heterogeneity of comorbidity networks increases over time in colorectal cancer pathways. (c) Onset of a comorbidity modulates future pathways.}
\label{fig:awesome_image3}
\end{figure*}
Heterogeneity of disease pathways among patients can be quantified by measuring the distance between their comorbidity networks. A commonly used metric for measuring distance between graphs is the Jaccard index as illustrated in Figure \ref{fig:awesome_image3} (a) \textcolor{blue}{\citep{real1996probabilistic}}. To handle weighted graphs, we use the weighted Jaccard index $J(X, Y) = \frac{\sum _i\min(x_i,y_i)}{\sum _i\max(x_i,y_i)}$ where $x_i$ and $y_i$ are the edge weights in two graphs X and Y.
Within a population, the average Jaccard distance $J_{avg}$ between any two individual networks measures the \textit{heterogeneity} of the population, i.e., 
\begin{equation}
    J_{avg} = \frac{\sum_{n<m}1-J(G_{n}, G_{m})}{{N \choose 2}}, 
\end{equation}
where $N$ denotes the size of population. The larger the average distance, the more spread out the population. Since the disease networks are time-varying, we compute $J_{avg}(t)$ based on the the networks at time $t$ to reflect the heterogeneity at that moment.

\paragraph{Patient pathways get more heterogeneous over time.}
In Figure \ref{fig:awesome_image3} (b), we track $J_{avg}$ over time (referenced to the date of initial diagnosis). We can readily see that the comorbidity networks learned by DDP become increasingly heterogeneous as time progresses. This reflects the fact that as a patient gets older, more comorbidities will occur and the subsequent disease pathway will become more complex. The increase in heterogeneity also highlights the need for modeling comorbidities with a personalized method since any one-size-fits-all approach will under-appreciate the diversity occurring later in the pathway. The ability of DDP to accurately predict the heterogeneity of colorectal cancer pathways is assessed in Section 4.3.

Figure \ref{fig:awesome_image3} (c) shows how the network dynamically adapts to ``influencers'', diseases which trigger a large variety of comorbidities and complications. The figure displays change in $J_{avg}$ before and after a disease onset relative to the population average. Positive value means increase in heterogeneity relative to the population, negative value otherwise, and zero means no change. Time is normalized so that the disease of interest always occurs at time 0. 
The red, blue and orange lines represent chronic obstructive pulmonary disease (COPD), Type 2 diabetes mellitus and heart failure respectively. It is well-established in the medical literature that all three types of diseases have complex heterogeneous comorbidity pathway \textcolor{blue}{\citep{fabbri2008complex, stratton2000association}}. This is clearly reflected in Figure \ref{fig:awesome_image3} (c) where the onset of these diseases triggers an immediate and persistent increase in the heterogeneity of comorbidity networks. On the other hand, the green line represents varicose veins of lower extremities, a mild condition that often does not need treatment \textcolor{blue}{\citep{vein}}. It is thus unsurprising to see that patients with this condition usually have less heterogeneous disease networks than the average. 

The above analysis shows DDP's ability to adjust the subsequent comorbidity pathway based on the occurrence of individual diseases. This high resolution view can help medical researchers better understand the progression and taxonomy of diseases.

\begin{figure*}[ht]
\begin{center}
\includegraphics[width=6.5in]{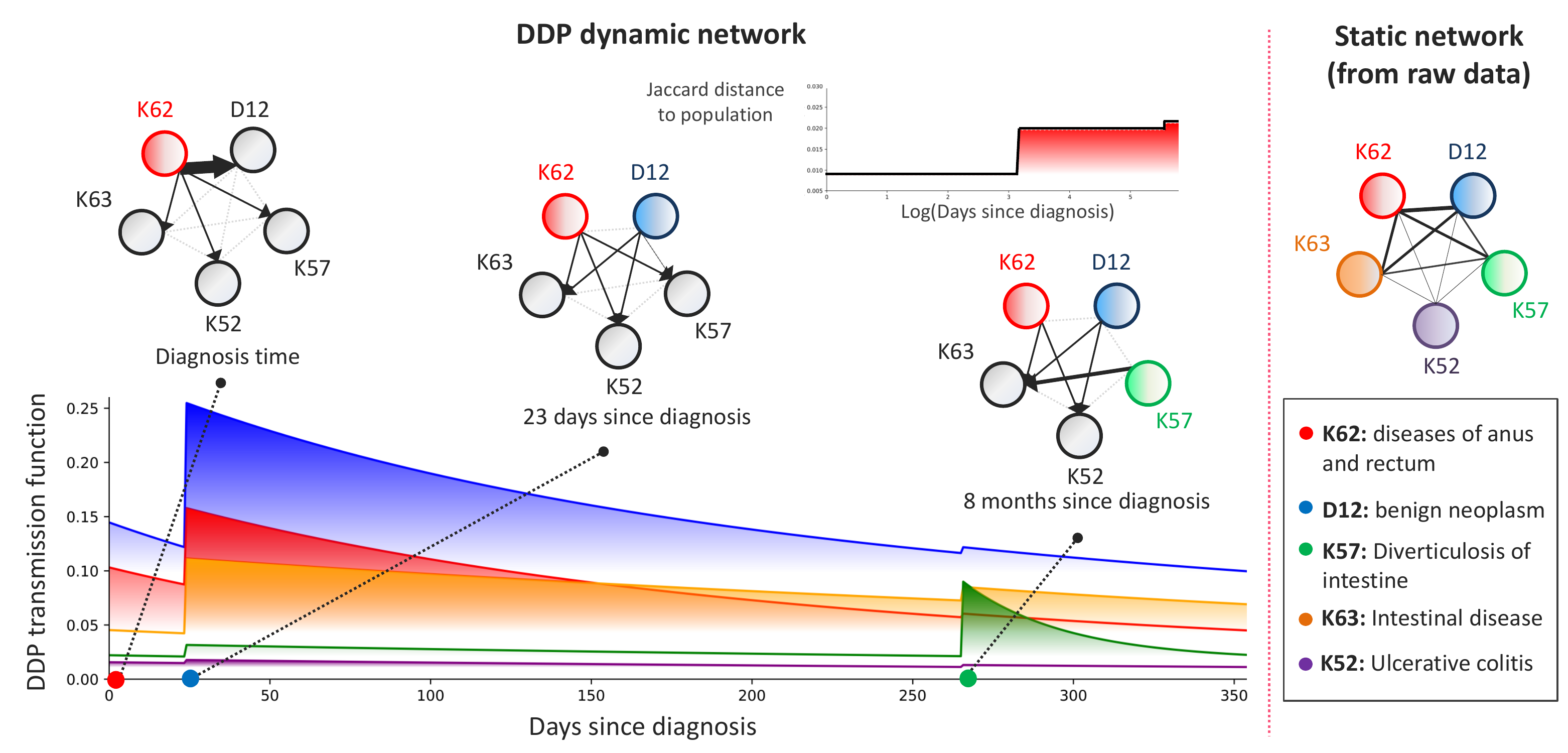}
\end{center}
\caption{The dynamic comorbidity network learned by DDP for an individual patient at three time steps, together with the corresponding intensity function. Nodes for diseases that have not occurred are colored in gray, and disease already diagnosed are assigned a distinct color. Edge thickness correspond to the disease likelihood at the given time step. In the upper left panel, we plot the Jaccard distance of the patient's network with respect to the average population as a function of time (on a logarithmic scale). The static comorbidity network obtained by counting disease co-occurrences and using the counts as graph edges is depicted on the right panel.}
\vspace{-.15in}
\label{fig:dyna_graph}
\end{figure*}

{\bf Individual-level comorbidity networks.} Figure \ref{fig:dyna_graph} depicts the evolution of the dynamic comorbidity network of five common gastrointestinal disorders (for one patient's pathway) as inferred by DDP --- the comorbidities included: diverticular diseases (ICD-10 code K57), intestinal disorders (K63), benign neoplasm in the colon and rectum (D12), diverticular diseases (K57), and ulcerative colitis (K52). The intensity function corresponding to the patient's trajectory is shown in the bottom panel. Each edge's thickness in the network corresponds to the likelihood of the disease designating the receiving node to occur at a given time step. The individual patient's dynamic network is contrasted with a static network (upper right panel) constructed directly from raw data by counting the co-occurrences of each pair of comorbidities and weighting edges accordingly. 

As we can see in Figure 4, the DDP comorbidity network is fairly dense at each time step, which suggests that the diseases are related. In fact, numerous medical publications have examined associations between these diseases. For example, it has been established that a lack of dietary fiber intake underlies the onset of diverticular diseases, intestinal disorders, and tumours of the colon and rectum \textcolor{blue}{\citep{painter1971diverticular, burkitt1972effect}}. Furthermore, there are strong epidemiological evidences of associations between tumours of colon and rectum, diverticular diseases, and ulcerative colitis \textcolor{blue}{\citep{ekbom1990ulcerative, burkitt1971epidemiology}}. 

Figure 4 shows that the dynamics of the inferred individualized comorbidity network cannot be deduced from the approach presented in \textcolor{blue}{\citet{beck2016diagnosis}}. That is, at each time step the RNN component of the DDP adapts the weights on the network edges to reflect the impact of previous diagnoses on the odds of future ones. Moreover, the weights of the network edges reflect the {\it timing} at which future comorbidities are expected to occur --- for instance, D12 occurs only 23 days after diagnosis, which was correctly anticipated by the DDP network as it assigned a large weight to the edge connecting the (pre-existing comorbidity) K62 and D12 at diagnosis time. On the contrary, K57 occured more than 8 months after diagnosis, which also was anticipated by DDP model having assigned a smaller weight to edges flowing into K57 node.   

By measuring the average (Jaccard) distance between the comorbidity network of the patient at hand and those of the overall patient population (upper left panel of Figure 4), we can see that the patient's network diverges from the typical population-level pathway as time progresses. This emphasizes the importance of the personalization aspect of the DDP model in predicting patient prognosis in later stages of the disease as we will show in the next Section.

\subsection{Predicting colorectal cancer pathways}
\label{sec:exp_tables}
\textbf{Prediction targets and evaluation metric.} Each individual patient has a unique disease pathway and a good representation of their health trajectory should enable differentiating these pathways. We evaluate how well DDP discerns the future disease pathways by predicting the next event. 
Given a disease history $\mathcal H_t = \{(t_i,v_i):t_i<t\}$, the models try to predict the probability of having a disease $v_{i+1}$ at time $t_{i+1}$. The time $t_{i+1}$ represents the time of the next disease onset available in the dataset. We chose to predict the incidental risk at a given time due to the nature of our data. For most chronic diseases including cancer, the diagnosis may occur much later than the actual onset. Hence the true disease onset time as well as the time between disease onsets are never observed. By focusing on predicting the diseases at known diagnosis time, evaluation becomes more objective and less prone to unknown variation. We calculate the Area Under ROC (AUC) score for predicting prevalent comorbidities. 

\begin{table*}[t]
\caption{{\footnotesize AUC ($\pm$ 95\% confidence intervals) performance for all baselines. Best performance is highlighted in bold font.}}
\label{tab:pred_res_high_degree}
\centering
{\footnotesize
\begin{tabular}{@{}lccccc@{}}
\toprule
\textbf{ICD-10 code} & \textbf{DDP} & \textbf{Neural Hawkes} & \textbf{cHawkes} & \textbf{Charlson} & \textbf{RETAIN} \\ \midrule
I50                                        & \textbf{0.74 $\pm$ 0.0114}                      & 0.72 $\pm$ 0.0127                                & 0.69 $\pm$ 0.0136                          & 0.68 $\pm$ 0.0111                           & \textbf{0.73 $\pm$ 0.0123}                   \\
N39  & \textbf{0.64 $\pm$ 0.0085}                      & 0.62 $\pm$ 0.0085                                & 0.59 $\pm$ 0.0083                          & 0.58 $\pm$ 0.0079                           & \textbf{0.65 $\pm$ 0.0085}                   \\
A41 & \textbf{0.72 $\pm$ 0.0091}                      & 0.72 $\pm$ 0.0092                                & 0.71 $\pm$ 0.0091                          & 0.60 $\pm$ 0.0098                           & 0.70 $\pm$ 0.0101                   \\
D12 & \textbf{0.69 $\pm$ 0.0053}                      & 0.67 $\pm$ 0.0055                                & 0.66 $\pm$ 0.0055                          & 0.59 $\pm$ 0.0055                           & 0.66 $\pm$ 0.0059                   \\
E86  & \textbf{0.72 $\pm$ 0.0225}                      & \textbf{0.72 $\pm$ 0.0235}                & 0.69 $\pm$ 0.0222                          & 0.52 $\pm$ 0.0222                           & 0.58 $\pm$ 0.0222                   \\
I25  & \textbf{0.79 $\pm$ 0.0081}                      & 0.77 $\pm$ 0.0089                                & 0.77 $\pm$ 0.0087                          & 0.63 $\pm$ 0.0085                           & 0.77 $\pm$ 0.0084                   \\
K63  & \textbf{0.68 $\pm$ 0.0061}                      & 0.64 $\pm$ 0.0064                                & 0.64 $\pm$ 0.0063                          & 0.60 $\pm$ 0.0060                           & 0.65 $\pm$ 0.0065                   \\
K83  & \textbf{0.69 $\pm$ 0.0217}                      & 0.68 $\pm$ 0.0225                                & 0.66 $\pm$ 0.0209                          & 0.63 $\pm$ 0.0200                           & 0.62 $\pm$ 0.0224                   \\  \bottomrule
\end{tabular}}
\end{table*}

\begin{table*}[t]
\caption{{\footnotesize Out-of-domain AUC performance for all baselines. Best performance is highlighted in bold font.}}
\label{tab:pred_res_stomach}
\centering
{\footnotesize
\begin{tabular}{@{}lccccc@{}}
\toprule
\textbf{ICD-10 code} & \textbf{DDP} & \textbf{Neural Hawkes} & \textbf{cHawkes} & \textbf{Charlson} & \textbf{RETAIN} \\ \midrule
I50                                        & \textbf{0.73 $\pm$ 0.0089} & 0.69 $\pm$ 0.0091                                & 0.65 $\pm$ 0.0102                          & 0.67 $\pm$ 0.0195                           & 0.71 $\pm$ 0.0093                   \\
N39                                        & \textbf{0.65 $\pm$ 0.0063} & 0.56 $\pm$ 0.0066                                & 0.59 $\pm$ 0.0064                          & 0.62 $\pm$ 0.0142                           & 0.63 $\pm$ 0.0066                   \\
A41                                        & \textbf{0.69 $\pm$ 0.0065} & 0.59 $\pm$ 0.0070                                & 0.65 $\pm$ 0.0070                          & 0.62 $\pm$ 0.0142                           & 0.66 $\pm$ 0.0072                   \\
D12                                        & \textbf{0.68 $\pm$ 0.0065} & 0.56 $\pm$ 0.0068                                & 0.66 $\pm$ 0.0065                          & 0.57 $\pm$ 0.0147                           & 0.63 $\pm$ 0.0078                   \\
E86                                        & \textbf{0.65 $\pm$ 0.0127} & 0.52 $\pm$ 0.0125                                & 0.62 $\pm$ 0.0121                          & 0.55 $\pm$ 0.0321                           & 0.56 $\pm$ 0.0122                   \\
I25                                        & \textbf{0.78 $\pm$ 0.0049} & 0.66 $\pm$ 0.0058                                & 0.75 $\pm$ 0.0054                          & 0.59 $\pm$ 0.0117                           & 0.75 $\pm$ 0.0053                   \\
K63                                        & \textbf{0.65 $\pm$ 0.0077} & 0.58 $\pm$ 0.0079                                & 0.60 $\pm$ 0.0078                          & 0.57 $\pm$ 0.0164                           & 0.63 $\pm$ 0.0083                   \\
K83                                        & \textbf{0.69 $\pm$ 0.0126} & 0.60 $\pm$ 0.0135                                & 0.65 $\pm$ 0.0123                          & 0.57 $\pm$ 0.0284                           & 0.63 $\pm$ 0.0128                   \\  \bottomrule
\end{tabular}}
\end{table*}

\textbf{Benchmarks.} We compare the performance of DDP with \textit{Neural Hawkes} \textcolor{blue}{\citep{mei2017neural}}, \textit{cHawkes} \textcolor{blue}{\citep{choi2015constructing}}, \textit{Charlson Score} \textcolor{blue}{\citep{charlson1987new}}, and RETAIN \textcolor{blue}{\citep{choi2016retain}}, which is a recurrent neural network with temporal attention mechanism akin to RPPN \textcolor{blue}{\citep{xiao2019learning}}. 
Section \ref{sec:related_work} contains a detailed review of these models.

\textbf{Results.} The results are shown in Table \ref{tab:pred_res_high_degree}. In five out of eight cases DDP achieved the best performance. In the rest three cases, the performance of DDP is comparable to that of Neural Hawkes or RETAIN. However, in these three cases, DDP does not only offer a competitive predictive accuracy, but infers the comorbidity network as well --- comorbidity networks cannot be straightforwardly inferred from the parameters of Neural Hawkes and RETAIN. This does not only provide more elaborate interpretability, but as we show in Section 4.4, it enables better generalization to out-of-domain data as we will. In addition, we can readily see that DDP always outperforms cHawkes and the Charlson Score. This suggests that the history-independent triggering mechanics of cHawkes and Charlson score do not adequately capture the disease complexity.

\subsection{Transferability to other types of cancer}
Finally, we applied all baselines originally trained on the primary dataset (colorectal cancer) to the out-of-domain dataset (stomach cancer) \textit{without} re-training. All other aspects of the experimental setup remains the same as the previous Section. The results are illustrated in Table \ref{tab:pred_res_stomach}. We can clearly see that DDP outperforms all the benchmarks including Neural Hawkes by a big margin on the out-of-domain samples. 

We performed a post-hoc analysis to better understand what the DDP model has learned. 
First, we performed Chi-squared tests to test whether the prevalence of individual diseases or the occurrence of disease pairs are different in the two data sets. In both cases, the test concluded that the distributions are different.($p$-value $<$ 0.001). This finding suggests that the disease networks constructed based on population level statistics such as those reviewed in Section \ref{sec:med_network} will tend to be \textbf{different} for the two datasets.  
Next, we randomly sampled a subset of patients from each of the two datasets and calculated Jaccard distance within and between the datasets. 
Student's t-test concluded the average distance between two groups are smaller than the distance within the group ($p$-value $<$ 0.01).
This indicates that the graph heterogeneity across datasets are smaller than the one within. In other words, the disease network learned by DDP applies to both sets of patients and is generalizable across cancer sites. 

\section{CONCLUSION}
In this paper, we developed DDP that utilizes deep neural networks to enable both {\it accurate} prediction of disease trajectories and {\it interpretable} representations of disease pathways. Combining the findings in Section 4, we can see that DDP can offer more nuanced understanding of disease progression mechanisms, more accurate prediction of patient pathways, and better generalizability across different diseases. By taking into account the full disease history, the learned DDP comorbidity networks are well equipped to deal with individual-level disease trajectories in a data-driven fashion, improving over existing one-size-fits-all clinical guidelines. The DDP model transferability is a major advantage in medical applications where the data for certain sub-populations are still scarce (e.g. in rare diseases). Moreover, the comorbidity networks uncovered by DDP may enable researchers to formulate hypotheses about the causal relations between diseases. 

\section{ACKNOWLEDGEMENTS}
The support and advice of the analytical and registration staff in Public Health England’s National Cancer Registration Service.  Only completely anonymous data was used in this work.



\bibliography{reference}
\bibliographystyle{conference}

\end{document}